\documentclass[10pt,twocolumn,letterpaper]{article}

\usepackage{cvpr}              

\usepackage[dvipsnames]{xcolor}

\usepackage{algorithm2e}
\usepackage{adjustbox}
\usepackage{multicol}
\usepackage{multirow}
\usepackage[dvipsnames]{xcolor}
\usepackage{booktabs}
\usepackage{color, colortbl}
\usepackage{pifont}

\newcommand{\ieno}{{\it i}.{\it e}.}
\newcommand{\egno}{{\it e}.{\it g}.}

\definecolor{Gray}{gray}{0.9}
\definecolor{myblue}{RGB}{80, 114, 153}
\definecolor{mygreen}{RGB}{46, 139, 87}
\definecolor{mypurple}{RGB}{153, 51, 204} 

\definecolor{cvprblue}{rgb}{0.21,0.49,0.74}
\usepackage[pagebackref,breaklinks,colorlinks,citecolor=cvprblue]{hyperref}

\title{Evolving Knowledge Mining for Class Incremental Segmentation}

\author{Zhihe Lu$^1$, Shuicheng Yan$^{2,3}$ and Xinchao Wang$^1$\\
$^1$National University of Singapore\\
$^2$Kunlun 2050 Research\\
$^3$Skywork AI\\
}

\begin{document}
\maketitle
\begin{abstract}
Class Incremental Semantic Segmentation (CISS) has been a trend recently due to its great significance in real-world applications. 
Although the existing CISS methods demonstrate remarkable performance, they either leverage the high-level knowledge (feature) only while neglecting the rich and diverse knowledge in the low-level features, leading to poor old knowledge preservation and weak new knowledge exploration; or use multi-level features for knowledge distillation by retraining a heavy backbone, which is computationally intensive. 
In this paper, we for the first time investigate the efficient multi-grained knowledge reuse for CISS, and propose a novel method, Evolving kNowleDge minING (ENDING), employing a frozen backbone.
ENDING incorporates two key modules: evolving fusion and semantic enhancement, for dynamic and comprehensive exploration of multi-grained knowledge.
Evolving fusion is tailored to extract knowledge from individual low-level feature using a personalized lightweight network, which is generated from a meta-net, taking the high-level feature as input. 
This design enables the evolution of knowledge mining and fusing when applied to incremental new classes.
In contrast, semantic enhancement is specifically crafted to aggregate prototype-based semantics from multi-level features, contributing to an enhanced representation.
We evaluate our method on two widely used benchmarks and consistently demonstrate new state-of-the-art performance.
The code is available at \href{https://github.com/zhiheLu/ENDING_ISS}{https://github.com/zhiheLu/ENDING\_ISS}.
\end{abstract}    
\section{Introduction}
\label{sec:intro}

\begin{figure}[ht]
    \centering
    \includegraphics[width=0.48\textwidth]{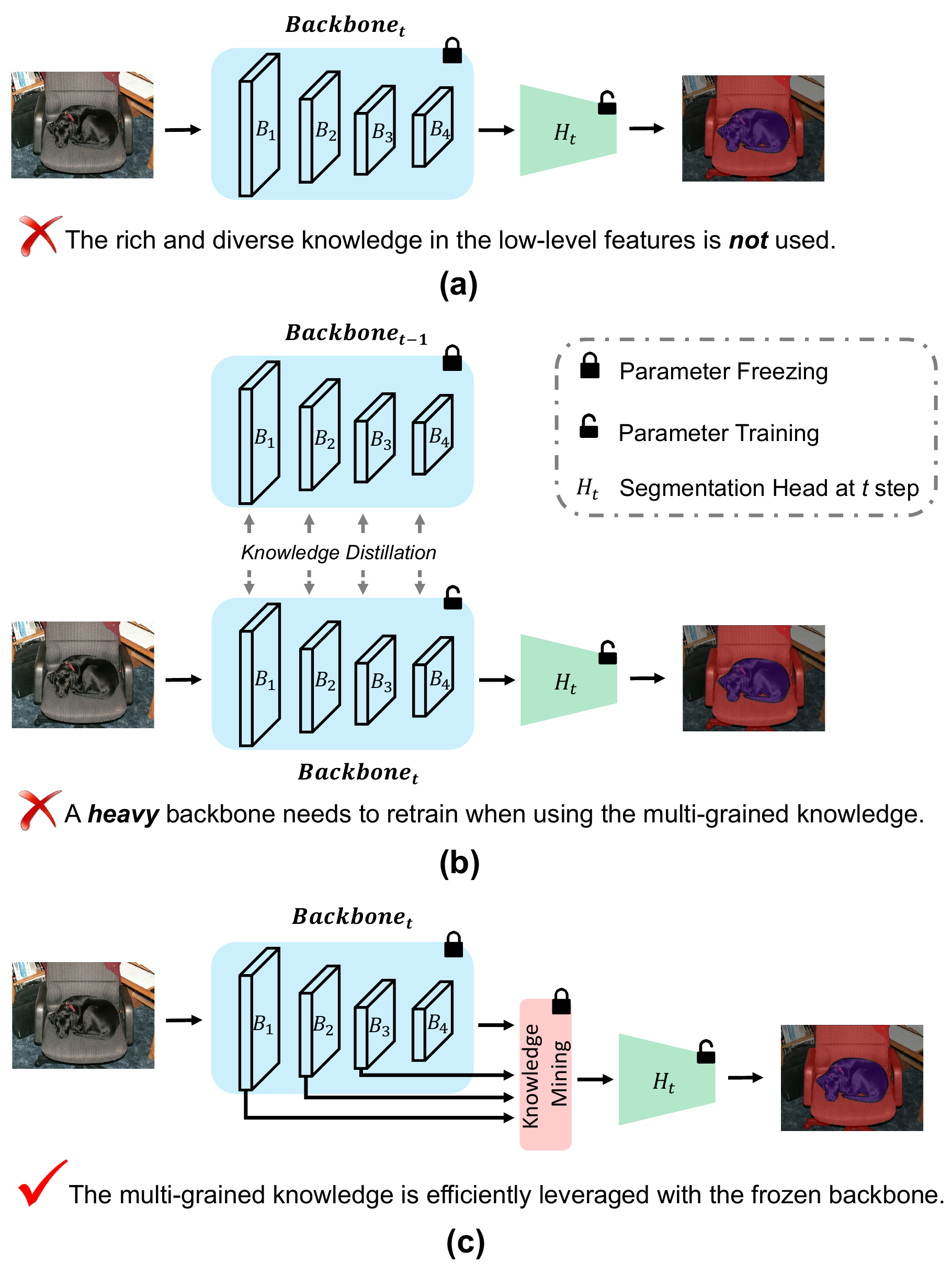}
    \caption{Illustration of the difference between our method {\bf (c)} and previous ones: {\bf (a)} and {\bf (b)}. 
    Our method is more advantageous by efficiently mining and reusing the multi-grained knowledge in varying-level features without retraining the heavy backbone.}
    \label{fig:illustration_diff}
\end{figure}

Deep models have achieved great success on semantic segmentation \cite{long2015fully,zhao2017pyramid,chen2018encoder,chen2017deeplab,chen2017rethinking,fu2019dual,huang2019ccnet,wang2018non,yuan2021ocnet,strudel2021segmenter,zheng2021rethinking} in a close-set fashion since 2015 \cite{long2015fully}, but those models inevitably faced catastrophic forgetting \cite{cauwenberghs2000incremental, mccloskey1989catastrophic,polikar2001learn++} -- a problem that was first proposed in 1989, \ieno, a model tends to forget the old knowledge when required to learn the new in sequential learning.
Solving this problem is crucial and practical, \egno, a self-driving system is always needed to recognize new environments or traffic signs without forgetting the old concepts.
Incremental learning \cite{li2017learning,hinton2015distilling,aljundi2017expert,mallya2018packnet,bang2021rainbow,belouadah2019il2m,chaudhry2021using,kim2021continual} provides a solution for this problem by learning a model in a continuous data stream, in which the learned model needs to perform well on both old and new scenarios.

\begin{figure}[ht]
    \centering
    \includegraphics[width=0.48\textwidth]{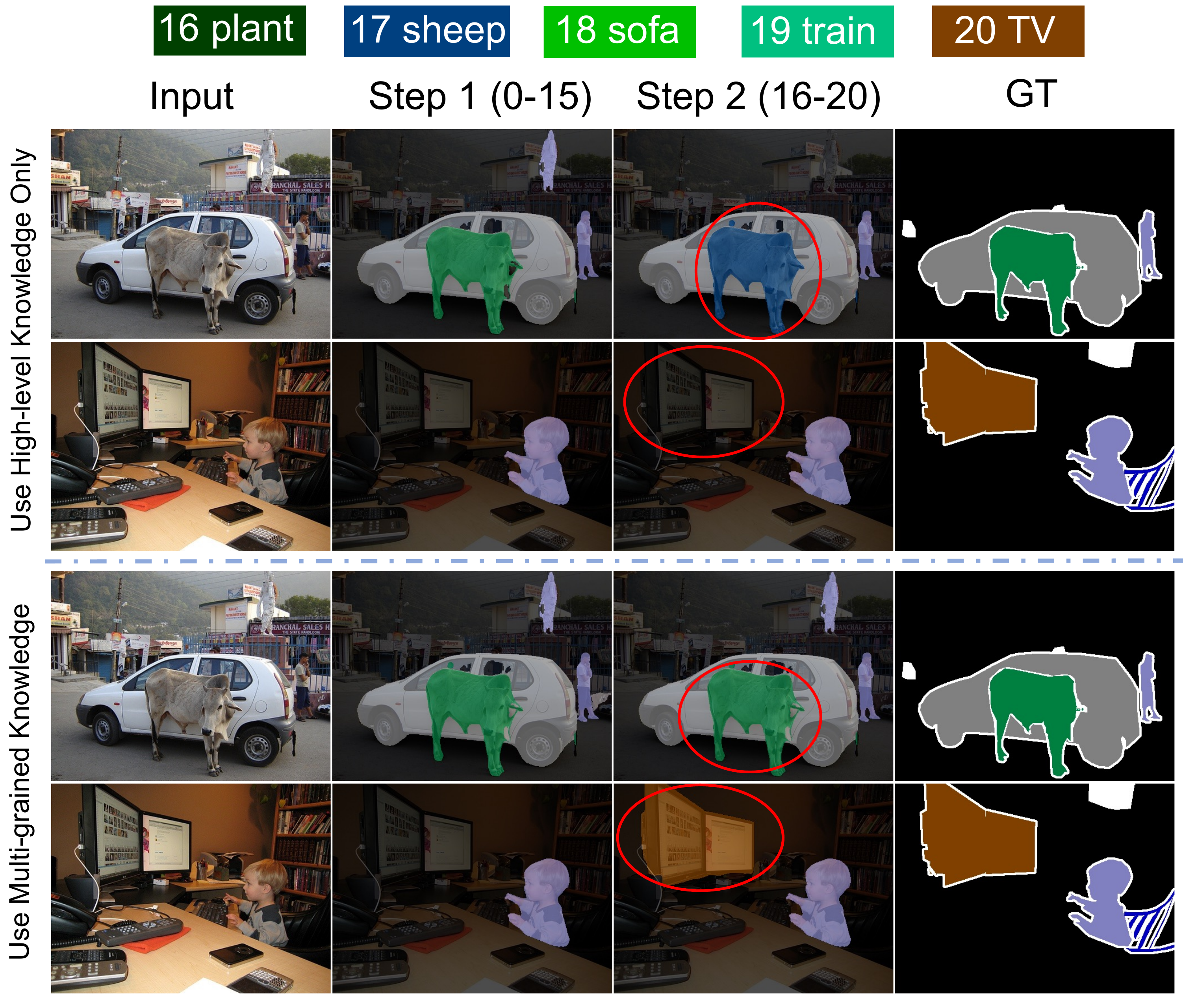}
    \caption{Illustrating the significance of using multi-grained knowledge for CISS. The top two rows are the results using the high-level knowledge (the feature $f_4$) only while the bottom two show the results with multi-grained knowledge (four-level features $f_1, f_2, f_3, f_4$) being used. $f_i$: the feature extracted from the $i$-th block of ResNet-101. With the high-level knowledge only, the model tends to neither forget the old knowledge learned in the first step, \egno, the ``cow'' is misclassified as ``sheep'' in the $1$-st row, nor lose the ability to learn new classes, \egno, the ``TV'' is totally missed ($2$-nd row). In contrast, using multi-grained knowledge (bottom two) can enhance both old knowledge preservation and new knowledge exploration for CISS. Baseline: MicroSeg \cite{zhang2022mining}. Set-up: {\bf 15-5 (2 steps)} on PASCAL VOC 2012.}
    \label{fig:illustration}
\end{figure}

Incremental learning has been introduced to semantic segmentation since 2019 \cite{michieli2019incremental}, but the historical works were specialized to classification problem, which are not suitable for class incremental semantic segmentation (CISS) due to its unique issue -- {\it background shift} \cite{cermelli2020modeling}.
That is, the background in step $t$ may contain not only current classes but also the past and future ones.
Regarding {\it background shift}, two effective techniques have been proposed: old knowledge based pseudo labeling \cite{douillard2021plop,cha2021ssul,zhang2022mining} and unknown class modeling \cite{cha2021ssul,zhang2022mining}.
Specifically, the pseudo labeling combines the pseudo labels from the model of step $t-1$ and the ground-truth labels at step $t$ to produce the updated labels containing both old and current classes, but it neglects the future classes that may exist in the background.
This limitation can be tackled by unknown class modeling, which often utilizes a class-agnostic model \cite{hou2017deeply,cheng2022masked} to detect all objects in one image.
These objects are then be used to model future classes.
However, those methods encounter two limitations: (i) only the high-level knowledge ({\it the output of the feature extraction backbone}) is leveraged to perform pseudo labeling and unknown class modeling \cite{cha2021ssul,zhang2022mining} while the low-level features that also contain abundant prior knowledge as well as more detailed semantics \cite{chen2018encoder} are not used as shown in Figure \ref{fig:illustration_diff} (a), weakening its capability to preserve old knowledge and learn new classes (see the first two rows in Figure \ref{fig:illustration}); 
and (ii) a heavy backbone, \egno, ResNet-101, is needed to retrain in the incremental step \cite{douillard2021plop,zhang2022representation} when reusing multi-level features for knowledge distillation.
Facing above limitations, an intuitive question might be raised: {\it can we efficiently reuse the multi-grained knowledge in diverse features without retraining a heavy backbone for CISS?}

Regarding the question, we first propose a simple multi-level feature aggregation module called na\"ive feature pyramid (NFP) and apply it to a state-of-the-art method -- MicroSeg \cite{zhang2022mining} (see Sec. \ref{sec:motivation} for details).
This module is only trained in the $1$-st step and is used to fuse features at various levels in subsequent steps.
Experimental results indicate that fusing high- and low-level features from any layer can boost the performance while aggregating all layers yields a new SOTA performance (see Table \ref{tab:low_level_effect}).
Qualitative illustrations in Figure \ref{fig:illustration} further suggest that this fusion enables a better utilization of the prior knowledge to serve incremental learning tasks, \ieno, better old knowledge preservation and new knowledge exploration.
However, it is worth noting that NFP is trained solely on classes in the $1$-st step and may exhibit bias towards those classes, potentially impeding the effective utilization of multi-level features in an incremental setup.

To that end, we propose a novel method, evolving knowledge mining (ENDING), to efficiently mine and reuse the multi-grained knowledge for CISS.
Specifically, ENDING consists of two key modules: evolving fusion and semantic enhancement, aimed at facilitating dynamic and thorough exploration of multi-grained knowledge.
The evolving fusion module is tailored to extract knowledge from each low-level feature by utilizing a personalized lightweight network generated from a meta-net, where the high-level feature serves as input.
This unique design allows for the evolution of knowledge mining when applied to incremental new classes.
The acquired multi-grained knowledge is then integrated at the feature level.
In contrast, the semantic enhancement module is crafted to aggregate prototype-based semantics from multi-level features, working at the semantic level.
This aggregation contributes to an enriched and refined representation of the semantic content.
To validate the effectiveness of ENDING, we conduct extensive experiments on two widely used benchmarks, demonstrating its superior performance.

The contributions are summarized as follows:
\begin{itemize}
    \item To our knowledge, this is the first study to investigate reusing multi-grained knowledge efficiently by leveraging multi-level features without the need to retrain a heavy backbone for CISS.
    We empirically demonstrate that the utilization of multi-grained knowledge is essential for CISS. 
    \item We propose a novel method, evolving knowledge mining (ENDING), for efficient multi-grained knowledge mining and reusing, which is achieved by two key modules: evolving fusion and semantic enhancement.
    \item We conduct extensive experiments to evaluate the performance of the proposed ENDING, and the results demonstrate that it yields new state-of-the-art performance on two widely used benchmarks. 
\end{itemize}

\section{Related Work}
\label{sec:relate}
\subsection{Semantic Segmentation} 
Semantic segmentation \cite{long2015fully,zhao2017pyramid,chen2018encoder,chen2017deeplab,chen2017rethinking,fu2019dual,huang2019ccnet,wang2018non,yuan2021ocnet,strudel2021segmenter,zheng2021rethinking,hariharan2015hypercolumns,wang2020deep} is a form of pixel-level prediction task that clusters parts of an image together which belong to the same object class.
The introduction of Fully Convolutional Networks (FCN) \cite{long2015fully} in 2015 marked a significant breakthrough in the field of semantic segmentation, and since then, numerous follow-up works have been proposed to improve the performance by introducing novel architectures, \egno, Dilated Convolutions \cite{yu2015multi}, Pyramid Pooling Module \cite{zhao2017pyramid}, U-Net \cite{ronneberger2015u}, Atrous Spatial Pyramid Pooling (ASPP) \cite{chen2017rethinking, chen2017deeplab}, and HRNet \cite{wang2020deep}.
Generally, those methods all do per-pixel classification with a regular multi-class cross-entropy loss.
In contrast, recent works \cite{cheng2021per,cheng2022masked} reformulated semantic segmentation as a mask classification task, \ieno, predicting a set of binary masks, each of which is associated with a global class label.
Despite the high performance in a close-set fashion, all mentioned methods suffer from {\it catastrophic forgetting} \cite{cauwenberghs2000incremental,mccloskey1989catastrophic,polikar2001learn++} when learning on new classes sequentially.
This problem has raised the interest of researchers to develop semantic segmentation models that can learn incrementally.
In this paper, we also follow this line to solve the incremental semantic segmentation task.

\subsection{Class Incremental Learning} 
Class incremental learning (CIL) aims to learn a classification model with the number of classes increasing sequentially.
Existing methods can be mainly grouped into three categories: parameter isolation \cite{aljundi2017expert,mallya2018packnet,rosenfeld2018incremental,serra2018overcoming}, replay-based \cite{de2021continual,lopez2017gradient,rebuffi2017icarl,rolnick2019experience,shin2017continual,oh2022alife} and regularization-based methods \cite{aljundi2018memory,li2017learning,rannen2017encoder,baek2022decomposed,chaudhry2018riemannian,dhar2019learning,douillard2020podnet,rebuffi2017icarl,simon2021learning,kirkpatrick2017overcoming,iscen2020memory,liu2020more,pan2020continual,park2019continual,tao2020topology,yu2020semantic,zenke2017continual,liu2022liqa}.
Specifically, parameter isolation based approaches often assign independent parameters for each incremental task, but face an increasing amount of parameters with the number of tasks.
Replay-based methods tend to either build a memory bank that stores a few samples of old classes \cite{de2021continual,lopez2017gradient,rebuffi2017icarl,rolnick2019experience} or learn a network to generate samples from old classes \cite{shin2017continual}.
Those samples are further used to jointly train with new classes for preventing old knowledge forgetting.
In contrast, regularization-based methods mainly focus on transferring old knowledge learned on old classes to the new model by knowledge distillation \cite{hinton2015distilling} or designing novel loss functions to prevent historical knowledge forgetting \cite{park2019continual,tao2020topology,yu2020semantic,zenke2017continual}.
Note that the mentioned techniques are not mutually exclusive but complementary.
In this paper, we propose an incremental learning method for semantic segmentation instead of classification.

\subsection{Class Incremental Semantic Segmentation} 
Class incremental semantic segmentation (CISS) is a similar task to CIL but aims to learn a segmentation model incrementally.
Due to the similarity, the techniques in CIL can be leveraged by CISS, \egno, knowledge distillation is used in ILT \cite{michieli2019incremental}, MiB \cite{cermelli2020modeling}, SDR \cite{michieli2021continual}, PLOP \cite{douillard2021plop}, RCN \cite{zhang2022representation}, and DKD \cite{baek2022decomposed}; replay-based techniques are adopted by RECALL \cite{maracani2021recall}, SSUL-M \cite{cha2021ssul}, MicroSeg-M \cite{zhang2022mining}, and ALIFE \cite{oh2022alife}.
However, CISS has its specific problem -- background shift \cite{cermelli2020modeling}, \ieno, the background in the $t$-th step may contain old or future classes, which cannot be solved by CIL methods.
To that end, recent works have proposed two useful techniques: old knowledge based pseudo labeling \cite{douillard2021plop,cha2021ssul,zhang2022mining} and unknown class modeling \cite{cha2021ssul,zhang2022mining}.
This particular pseudo labeling complements the ground truth labels in the $t$-th step with pseudo labels of old classes predicted by the model of $(t-1)$-th step while unknown class modeling tends to utilize a pre-trained class-agnostic model for possible objects detection and future class modeling.
Despite the effectiveness, those methods exist some limitations: (i) the old knowledge in low-level features with much subtler details are overlooked when performing pseudo labeling and unknown class modeling \cite{cha2021ssul,zhang2022mining}; (ii) a heavy backbone is required to retrain in the incremental step \cite{douillard2021plop,zhang2022representation} when using multi-level features for knowledge distillation.
In this paper, we propose to efficiently leverage the multi-grained knowledge in various features without retraining the parameters of backbone.
\section{Methodology}

\begin{figure*}[ht]
    \centering
    \includegraphics[width=0.95\textwidth]{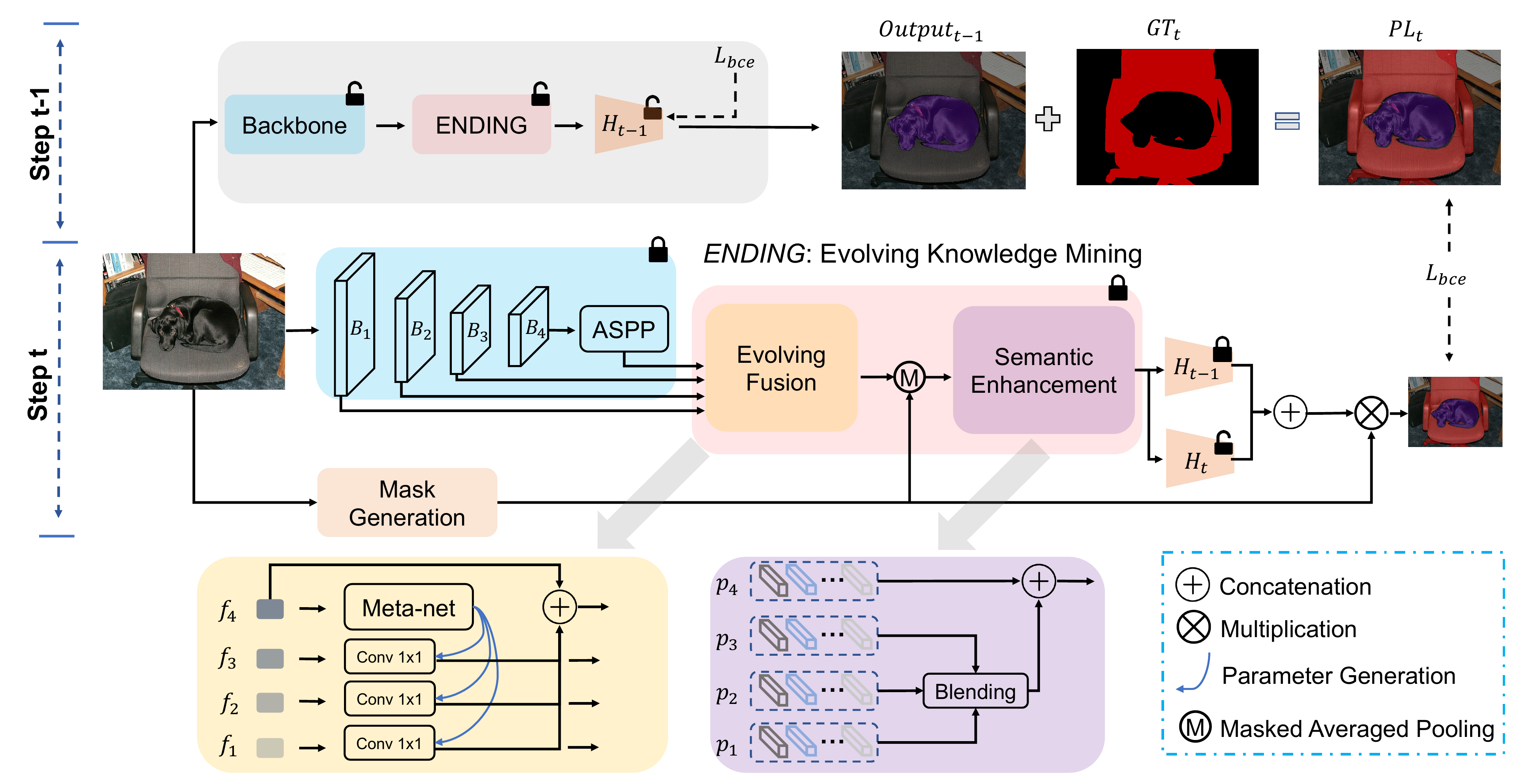}
    \caption{The overview of our evolving knowledge mining (ENDING). In the $1$-st step, the entire model including backbone, ENDING, and segmentation head $H_1$ are trained on $C_1$ classes. 
    During training, ENDING learns to mine the knowledge in low-level features through personalized networks generated by a meta-net, taking the high-level feature as the input.
    The mined knowledge is then fused at both feature and semantic (prototype) level.
    In the follow-up steps, only a new segmentation head $H_t$ is trainable for $C_t$ classes with our ENDING to dynamically mine the knowledge for those classes relying on the corresponding high-level feature. 
    $C_t$: the classes shown in step $t$.}
    \label{fig:framework}
\end{figure*}

\subsection{Problem Set-up}
Class incremental semantic segmentation (CISS) \cite{cha2021ssul,zhang2022mining} aims to learn a segmentation model incrementally by updating part/all of the parameters of the historical model and a set of new parameters specialized to the current task.
Specifically, one CISS task often consists of $T$ learning steps, each with a dataset $D_t$ from $C_t$ classes.
Note that $D_t$ may also contain historical classes from previous steps or future classes, but these classes are labeled as background, which results in a unique problem in CISS -- {\it background shift}.
Except background class, the given classes in the $t$-th step are disjoint with the classes from other steps, \ieno, $C_t \cap C_i = \emptyset$, where $i \in \{1, \dots, T\} \setminus \{t\}$.
In step $t$, a model is learned on $C_t$ classes only, the learned model tends to forget the historical knowledge acquired from old classes without accessing $C_1 \cup \dots \cup C_{t-1}$ classes, thus causing {\it catastrophic forgetting} -- the key problem in incremental learning.
Overall, an incremental segmentation model is designed to tackle both {\it background shift} and {\it catastrophic forgetting}, thereby achieving good pixel-wise recognition performance on the current classes as well as the classes from the past steps, \ieno, $C_1 \cup \dots \cup C_t$.

\subsection{Overview}
In general, a CISS model consists of a heavy backbone $F_b$ and $T$ light-weight segmentation heads $\{H_1, \dots, H_T\}$, in which the $F_b$ and $H_1$ are fully trained in the $1$-st step and then [$F_b$ and $H_t$] or [only $H_t$] will be updated in step $t$.
When learning the model in step $t$, it is common to borrow the old knowledge learned in step $t-1$ by using the features of that model, \egno, four-level features extracted from four blocks of ResNet-101 \cite{he2016deep} -- the standard backbone used in existing methods.
The weakness of current CISS methods is that they either mine the prior knowledge stored in the high-level features only \cite{cha2021ssul,zhang2022mining}; or explore more knowledge in multi-level features but retrain the heavy backbone $F_b$ \cite{douillard2021plop,zhang2022representation}.

To that end, this work focuses on a better utilization of prior knowledge by efficiently leveraging multi-level features for improved CISS by training $H_t$ only.
Specifically, we propose evolving knowledge mining (ENDING) for dynamic knowledge mining and fusing, thereby improving the performance on both old and new classes.
After a one-time training in the $1$-st step, our ENDING can effectively mine the multi-grained knowledge in low-level features, conditioned on the high-level feature; and then integrate the acquired knowledge at both the feature and semantic levels, facilitating subsequent incremental segmentation tasks.
ENDING comprises two key modules: evolving fusion and semantic enhancement, each designed for distinct purposes. 

\subsection{Evolving Fusion}
Evolving fusion is designed to dynamically extract knowledge from multiple low-level features, facilitating an enhanced knowledge aggregation, as shown in Figure \ref{fig:framework}.
The dynamic knowledge extraction is achieved by mining the multi-grained knowledge of low-level features using a personalized network.
This network is generated from a meta-net, taking the corresponding high-level feature as the input.
The multi-grained knowledge is then aggregated through concatenation at the feature level.
To be specific, we formulate the above process as follows:
\begin{align}
    \phi_1, \phi_2, \phi_3 &= \Theta_1(f_4), \Theta_2(f_4), \Theta_3(f_4) \label{eq:personalized} \\
    k_1, k_2, k_3 &= \phi_1(f_1), \phi_2(f_2), \phi_3(f_3) \label{eq:knowledge} \\
    f_{out} &= concat(f_4, k_1, k_2, k_3) \label{eq:fusion}
\end{align}
, where $f_i, i \in \{1, \dots, 4\}$ is the $i$-th level feature, $\Theta_i$ is a sub-net of meta-net for the network generation, $\phi_i$ is a personalized network for the $i$-th low-level feature, $k_i$ represents the ``knowledge'' feature extracted from the $i$-th low-level feature.

\paragraph{Remark}
Our evolving fusion is designed for dynamic adaptation to new classes, facilitating efficient knowledge mining and fusing. 
The key lies in the training of our meta-net, which is adept at using high-level features as input to generate personalized networks. 
These networks are well trained to extract knowledge from corresponding low-level features. 
The class-agnostic nature of this training enables adaptability to paired high- and low-level features of incremental new classes, ensuring effective knowledge mining and fusing in an evolving fashion.

\subsection{Semantic Enhancement}
In addition to mining and fusing multi-grained knowledge at the feature level, as demonstrated in our evolving fusion module, we augment the representation through knowledge aggregation at the semantic level by our semantic enhancement (illustrated in Figure \ref{fig:framework}). 
The semantic features are represented by prototypes, computed using masked average pooling. 
Specifically, considering both the output and the extracted low-level features from evolving fusion, we compute the prototypes as follows:
\begin{equation}
    p_1, p_2, p_3, p_4 = \mathcal{M}(k_1), \mathcal{M}(k_2), \mathcal{M}(k_3), \mathcal{M}(f_{out}) \label{eq:proto}
\end{equation}
, where $\mathcal{M}$ is the masked average pooling operation and $p_i$ is the prototype for the $i$-th level feature.

Then, three low-level prototypes $p_1, p_2, p_3$ are blended for further semantic enhancement, which is defined as:
\begin{align}
    p_b &= \theta(\sum_{i=1}^3 p_i) \label{eq:enhance_1} \\
    p_{out} &= concat(p_4, p_b) \label{eq:enhance_2}
\end{align}
, where $\theta$ is a two-layer MLP for prototype blending.

\subsection{Objective Function}
Following the work \cite{zhang2022mining}, we have two loss functions: $\mathcal{L}_{bce}$ and $\mathcal{L}_c$.
$\mathcal{L}_{bce}$ is the loss function employed on the outputs of two segmentation heads, taking $f_{out}$ and $p_{out}$ as the inputs, respectively.
The fomulation is defined as follows:
\begin{align}
    \mathcal{L}_{bce} &= \sum_{i=1}^{HW} BCE(\hat{y}_1^i, \widetilde{y}^i) + \sum_{i=1}^{HW} BCE(\hat{y}_2^i, \widetilde{y}^i) \label{eq: bce} \\
    \hat{y}_1 &= concat(\{h_{1}(f_{out}), \dots, h_t(f_{out})\}) \label{eq: y1}  \\
    \hat{y}_2 &= concat(\{h_{1}^{\star}(p_{out}), \dots, h_t^{\star}(p_{out})\}) \otimes mask \label{eq: y2} 
\end{align}
, where $BCE$ is the binary cross-entropy, $\widetilde{y}$ are the enhanced labels through combining both pseudo labels generated from the $(t-1)$-th model and the ground-truth label of the $t$-th step, $HW$ is the spatial dimension of the prediction, $\{h_{i}\}$ are the segmentation heads for $f_{out}$, $\{h_{i}^{\star}\}$ are for $p_{out}$, $mask$ is the concatenation of N binary masks and $\hat{y}_2$ is obtained by multiplication between the concatenated outputs ($\mathcal{R}\in N_{cls} \times N$) and mask ($\mathcal{R}\in N \times H \times W$). Note that the first part of $\mathcal{L}_{bce}$ is only used in the $1$-st step.

$\mathcal{L}_c$ is a contrastive loss \cite{he2020momentum} for regularizing unseen classes modeling, which is defined as:
\begin{equation}
    \mathcal{L}_c = -\frac{1}{K} \sum_{j=1}^K log \frac{exp(o_j \cdot o_j)}{\sum_{m=1}^K exp(o_j \cdot o_m)} \label{eq: unseen} 
\end{equation}
, where $o_j$ is the prediction of ``unseen'' cluster. Please refer to \cite{zhang2022mining} for more details.

The final objective function is the combination of the two losses, \ieno, $\mathcal{L} = \mathcal{L}_{bce} + \mathcal{L}_c$. 
For clarification, we provide the pseudo code in algorithm \ref{alg:train}.

\RestyleAlgo{ruled}
\SetKwComment{Comment}{$\triangleright$\ }{}

\begin{algorithm}[ht]
\small
\caption{Training process}\label{alg:train}
\KwData{Dataset $\{D_t, t \in \{1, \dots, T\}\}$, Backbone $\Omega$, Meta-net $\Theta$, Segmentation Heads $\{H_i\}$}
\KwResult{Updated $\Omega$, $\Theta$ and $\{H_i\}$}
 Model initialization and training hyper-parameter setup\;
 \For{$step=1,\dots,T$}{
    \BlankLine
    \textcolor{mygreen}{{\bf // Freeze Parameters}}\;
    \If{$step>1$}{Freeze $\Omega$ and $\Theta$\;}
    \BlankLine
    Extract features $f_i, i \in \{1, \dots, 4\}$ from $\Omega(x), x \in D_t$\;
    
    \BlankLine
    \textcolor{myblue}{{\bf // Evolving Fusion}}\;
    Personalized network generation as per Eq. \ref{eq:personalized}\;
    Knowledge mining following Eq. \ref{eq:knowledge}\;
    Obtain $f_{out}$ from evolving aggregation (Eq. \ref{eq:fusion});

    \BlankLine
    \textcolor{mypurple}{{\bf // Semantic Enhancement}}\;
    Compute prototypes as per Eq. \ref{eq:proto}\;
    Knowledge blending (Eq. \ref{eq:enhance_1})\;
    Enhance semantic following Eq. \ref{eq:enhance_2}\;

    \BlankLine
    \textcolor{mygreen}{{\bf // Optimization}}\;
    \eIf{$step>1$}{
       Obtain predictions $\hat{y_2}$ via $\{H_i\}$ (Eq. \ref{eq: y2})\;
       Compute loss $\mathcal{L}_{bce}+\mathcal{L}_c$\; \Comment{Only the second part of Eq.\ref{eq: bce} is used for $\mathcal{L}_{bce}$}
       Compute gradients and update parameters\;
       }{
       Obtain predictions $\hat{y_1}, \hat{y_2}$ via $\{H_i\}$ (Eq. \ref{eq: y1}, \ref{eq: y2})\;
       Compute loss $\mathcal{L}_{bce}+\mathcal{L}_c$\;
       Compute gradients and update parameters\;
      }
 }
\end{algorithm}
\section{Experiments}

\subsection{Datasets}
\paragraph{PASCAL VOC 2012 \cite{everingham2009pascal}} It is a widely used segmentation dataset, which can be repurposed for CISS.
It contains 10,582/1,449 training/validation images from 21 classes including 20 object classes and one background class.
We follow the previous works \cite{cha2021ssul,zhang2022mining,zhang2022representation} to conduct experiments on multiple incremental tasks as shown in Table \ref{tab:voc_main}.
For example, in the {\bf 10-1 (11-steps)} task, a model is trained on 11 classes (including the background class) in the $1$-st step, and then sequentially trained on one new class from step 2 to step 11.
Note that all experiments are under the more realistic {\it overlapped} set-up as per \cite{cha2021ssul,zhang2022mining,zhang2022representation,douillard2021plop}, \ieno, the background of an image in step $t$ may contain old or future classes while only the new classes belonging to the $t$-th step are labeled, except for Sec. \ref{sec:disjoint} that evaluates our method in the {\it disjoint} set-up, which assumes all future classes are known and {\it not} shown in the background.

\paragraph{ADE20K \cite{zhou2017scene}} This is a more challenging semantic segmentation dataset with 150 classes for daily life scenes, which contains 20,210/2,000 training/validation images.
The experimental protocol for ADE20K is in line with that of PASCAL VOC 2012.

\subsection{Implementation Details}\label{sec:implement}
Following prior works \cite{cermelli2020modeling, cha2021ssul, zhang2022mining}, we adopt DeepLabV3 \cite{chen2017deeplab} with a ResNet-101 backbone, which is pre-trained on ImageNet \cite{deng2009imagenet}.
During training, we use SGD optimizer with the momentum 0.9 and weight decay 1e-4, and use the initial learning rate 1e-2 and batch size 16 for PASCAL VOC 2012 in all steps, while 5e-3 ($1$-st step) / 3e-2 (other steps) and 12 for ADE20K.
We train the model for 50 epochs per step.
The learning rate is scheduled by ``poly'' learning rate policy \cite{chen2017deeplab}.
The data augmentations used here are random scaling (from 0.5 to 2.0) and random flipping.
As per \cite{zhang2022mining}, the mask generator is Mask2Former \cite{cheng2022masked} pre-trained on MS-COCO \cite{lin2014microsoft}, which produces 100 class-agnostic proposals for each image.
It's worth noting that the mask generator is not fine-tuned on
any benchmark dataset for a fair comparison. 
The hyper-parameters for unseen classes modeling are $K=5$ for PASCAL VOC 2012 and $K=1$ for ADE20K as per \cite{zhang2022mining}.
When using replaying, the samples of extra memory is 100 for PASCAL VOC 2012. 
Our experiments are implemented with PyTorch on two NVIDIA Tesla V100 GPUs.

\subsection{Compared Methods}
We compare existing methods for a comprehensive evaluation.
Specifically, three classical methods in CIL, \ieno, EWC \cite{kirkpatrick2017overcoming}, Lwf-MC (TPAMI2017) \cite{li2017learning}, and ILT (ICCVW2019) \cite{michieli2019incremental}, are repurposed for CISS and compared.
Moreover, we make comparison with the existing state-of-the-art CISS methods including MiB (CVPR2020) \cite{cermelli2020modeling}, SDR (CVPR2021) \cite{michieli2021continual}, PLOP (CVPR2021) \cite{douillard2021plop}, RCIL (CVPR2022) \cite{zhang2022representation}, SSUL (NeurIPS2021) \cite{cha2021ssul}, MicroSeg (NeurIPS2022) \cite{zhang2022mining}, and EWF (CVPR2023) \cite{xiao2023endpoints}.

\subsection{Evaluation Metric}
We evaluate our method using the mean Intersection-over-Union (mIoU) metric, which is computed as the average of the Intersection-over-Union (IoU) scores for each class. 
The IoU score for a class is computed as the true positives divided by the sum of true positives, false positives, and false negatives.
In all tables, we reports three mIoU metrics: the mIoU of the classes in the first step, the mIoU of the classes in the follow-up steps and the mIoU of all classes.
The first metric evaluates the model's ability to preserve old knowledge, while the second measures its ability to learn new classes. 
The final metric reflects the model's overall performance on both old and new classes.

\subsection{Motivation Experiments}
\label{sec:motivation}
We conduct the first experiment by incorporating multiple features into a baseline method -- MicroSeg \cite{zhang2022mining} by the na\"ive feature pyramid, as presented in Table \ref{tab:low_level_effect}.
Specifically, each low-level feature $f_i$ ($i\in\{1,2,3\}$) undergoes projection via a 1$\times$1 Conv layer to dimension $48 \times H \times W$. 
Subsequently, these features are concatenated with the high-level feature $f_4$ to serve as input for the subsequent segmentation heads. 
The results indicate that adding any low-level feature to $f_4$ leads to performance improvement, with the best performance achieved when utilizing all-level features. 
This suggests that leveraging multi-level features indeed enhances the model's capacity for incremental learning, providing motivation for the design of our evolving knowledge mining method to better suit incremental learning scenarios. 

\begin{table}[ht]
    \centering
    \adjustbox{max width=0.3\textwidth}{
    \begin{tabular}{ccccccc}
        \toprule
        \multirow{2}{*}{$f_1$} & \multirow{2}{*}{$f_2$} & \multirow{2}{*}{$f_3$} & \multirow{2}{*}{$f_4$} & \multicolumn{3}{c}{\textbf{15-5 (2 steps)}} \\
         & & &  & 0-15 & 16-20 & all \\
         \midrule
         & & & $\checkmark$ & 80.4 & 52.8 & 73.8 \\
         $\checkmark$ & & & $\checkmark$ & 81.5 & 50.8 & 74.2 \\
         & $\checkmark$ & & $\checkmark$ & 80.7 & 53.1 & 74.1 \\
         & & $\checkmark$ & $\checkmark$ & 80.6 & 53.2 & 74.1 \\
         $\checkmark$ & $\checkmark$ & & $\checkmark$ & 81.3 & 51.7 & 74.3 \\
         $\checkmark$ & & $\checkmark$ & $\checkmark$ & 81.4 & 52.3 & 74.5 \\
         & $\checkmark$ & $\checkmark$ & $\checkmark$ & 81.6 & 51.4 & 74.4 \\
         $\checkmark$ & $\checkmark$ & $\checkmark$ & $\checkmark$ & \textbf{81.7} & \textbf{53.7} & \textbf{75.0} \\
         \bottomrule
    \end{tabular}}
    \caption{The effect of leveraging multi-level features on the SOTA CISS method -- MicroSeg \cite{zhang2022mining}. 
    Architecture: DeepLabV3 \cite{chen2017deeplab} with the backbone ResNet-101 \cite{he2016deep}. $f_{i}, i\in\{1,\dots,4\}$ is the feature extracted from the $i$-th block of ResNet-101. 
    Setting: {\bf 15-5 (2 steps)} on PASCAL VOC 2012.}
    \label{tab:low_level_effect}
\end{table}

\begin{table*}[ht]
    \adjustbox{max width=\textwidth}{
    \centering
    \begin{tabular}{l|ccc|ccc|ccc|ccc|ccc|ccc}
         \toprule
         \multirow{2}{*}{Method} & \multicolumn{3}{c}{{\bf 10-1 (11 steps)}} & \multicolumn{3}{c}{{\bf 2-2 (10 steps)}} & \multicolumn{3}{c}{{\bf 15-1 (6 steps)}} & \multicolumn{3}{c}{{\bf 5-3 (6 steps)}} & \multicolumn{3}{c}{{\bf 19-1 (2 steps)}} & \multicolumn{3}{c}{{\bf 15-5 (2 steps)}} \\
         & 0-10 & 11-20 & all & 0-2 & 3-20 & all & 0-15 & 16-20 & all & 0-5 & 6-20 & all & 0-19 & 20 & all & 0-15 & 16-20 & all \\
         \midrule
         Joint (Upper bound) & 82.1 & 79.6 & 80.9 & 76.5 & 81.6 & 80.9 & 82.7 & 75.0 & 80.9 & 81.4 & 80.7 & 80.9 & 81.0 & 79.1 & 80.9 & 82.7 & 75.0 & 80.9 \\
         \midrule
         EWC \cite{kirkpatrick2017overcoming} (NAC17) & - & - & - & - & - & - & 0.3 & 4.3 & 1.3 & - & - & - & 26.9 & 14.0 & 26.3 & 24.3 & 35.5 & 27.1 \\
         LwF-MC \cite{li2017learning} (TPAMI17) & 4.7 & 5.9 & 4.9 & 3.5 & 4.7 & 4.5 & 6.4 & 8.4 & 6.9 & 20.9 & 36.7 & 24.7 & 64.4 & 13.3 & 61.9 & 58.1 & 35.0 & 52.3 \\
         ILT \cite{michieli2019incremental} (ICCVW19) & 7.2 & 3.7 & 5.5 & 5.8 & 5.0 & 5.1 & 8.8 & 8.0 & 8.6 & 22.5 & 31.7 & 29.0 & 67.8 & 10.9 & 65.1 & 67.1 & 39.2 & 60.5 \\
         MiB \cite{cermelli2020modeling} (CVPR20)   & 12.3 & 13.1 & 12.7 & 41.1 & 23.4 & 25.9 & 34.2 & 13.5 & 29.3 & 57.1 & 42.6 & 46.7 & 71.4 & 23.6 & 69.2 & 76.4 & 50.0 & 70.1 \\
         SDR \cite{michieli2021continual} (CVPR21)  & 32.1 & 17.0 & 24.9 & 13.0 & 5.1 & 6.2 & 44.7 & 21.8 & 39.2 & 12.1 & 6.5 & 8.1 & 69.1 & 32.6 & 67.4 & 57.4 & 52.6 & 69.9 \\
         PLOP \cite{douillard2021plop} (CVPR21)  & 44.0 & 15.5 & 30.5 & 24.1 & 11.9 & 13.7 & 65.1 & 21.1 & 54.6 & 17.5 & 19.2 & 18.7 & 75.4 & {\bf 37.4} & 73.5 & 75.7 & 51.7 & 70.1 \\
         SSUL \cite{cha2021ssul} (NeurIPS21)  & 71.3 & 46.0 & 59.3 & 62.4 & 42.5 & 45.3 & 77.3 & 36.6 & 67.6 & 72.4 & 50.7 & 56.9 & 77.7 & 29.7 & 75.4 & 77.8 & 50.1 & 71.2 \\
         RCIL \cite{zhang2022representation} (CVPR22)  & 55.4 & 15.1 & 34.3 & - & - & - & 70.6 & 23.7 & 59.4 & - & - & - & - & - & - & 78.8 & 52.0 & 72.4 \\
         MicroSeg \cite{zhang2022mining} (NeurIPS22)  & 72.6 & 48.7 & 61.2 & 61.4 & 40.6 & 43.5 & 80.1 & 36.8 & 69.8 & 77.6 & 59.0 & 64.3 & 78.8 & 14.0 & 75.7 & 80.4 & 52.8 & 73.8 \\
         EWF \cite{xiao2023endpoints} (CVPR23) & 71.5 & 30.3 & 51.9 & - & -& - & 77.7 & 32.7 & 67.0 & 61.7 & 42.2 & 47.7 & 77.9 & 6.7 & 74.5 & - & - & - \\
         \midrule
         \rowcolor{Gray}
         ENDING (Ours)  & {\bf 73.0} & {\bf 50.1} & {\bf 62.1} & {\bf 67.3} & {\bf 55.8} & {\bf 57.4} & {\bf 80.5} & {\bf 39.3} & {\bf 70.7} & {\bf 78.1} & {\bf 62.4} & {\bf 66.9} & {\bf 80.0} & 22.2 & {\bf 77.3} & {\bf 81.5} & {\bf 56.1} & {\bf 75.5} \\
         \bottomrule
    \end{tabular}}
    \caption{Quantitative comparison on PASCAL VOC 2012. The results indicate that our ENDING yields the state-of-the-art performance in most cases. The best performance is in {\bf bold}.}
    \label{tab:voc_main}
\end{table*}
\begin{figure*}[ht]
    \centering
    \includegraphics[width=\textwidth]{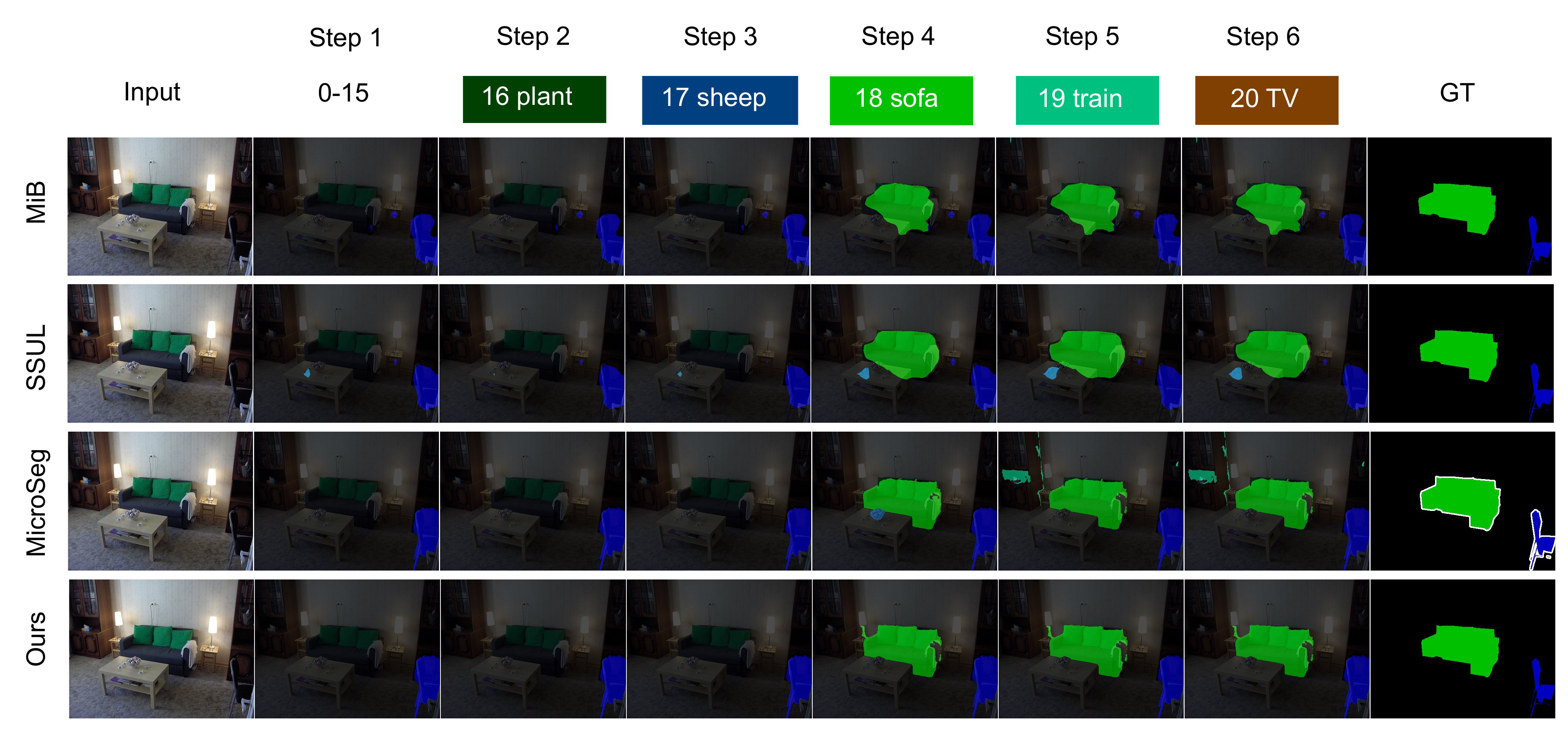}
    \caption{Qualitative comparison with baseline methods on {\bf 15-1 (6 steps)} setting on PASCAL VOC 2012. Zoom in for details.}
    \label{fig:visualization}
\end{figure*}

\begin{table*}[ht]
    \centering
    \adjustbox{max width=0.88\textwidth}{
    \begin{tabular}{l|ccc|ccc|ccc|ccc}
    \toprule
        \multirow{2}{*}{Method} & \multicolumn{3}{c}{{\bf 100-5 (11 steps)}} & \multicolumn{3}{c}{{\bf 100-10 (6 steps)}} & \multicolumn{3}{c}{{\bf 100-50 (2 steps)}} & \multicolumn{3}{c}{{\bf 50-50 (3 steps)}}  \\
        & 0-100 & 101-150 & all & 0-100 & 101-150 & all & 0-100 & 101-150 & all & 0-50 & 51-150 & all \\
        \midrule
        Joint (Upper bound) & 43.8 & 28.9 & 38.9 & 43.8 & 28.9 & 38.9 & 43.8 & 28.9 & 38.9 & 50.7 & 32.8 & 38.9 \\
        \midrule
        ILT \cite{michieli2019incremental} (ICCVW19)  & 0.1 & 1.3 & 0.5 & 0.1 & 3.1 & 1.1 & 18.3 & 14.4 & 17.0 & 3.5 & 12.9 & 9.7 \\
        MiB \cite{cermelli2020modeling} (CVPR20)  & 36.0 & 5.7 & 26.0 & 38.2 & 11.1 & 29.2 & 40.5 & 17.2 & 32.8 & 45.6 & 21.0 & 29.3 \\
        PLOP \cite{douillard2021plop} (CVPR21)  & 39.1 & 7.8 & 28.8 & 40.5 & 13.6 & 31.6 & 41.9 & 14.9 & 32.9 & 48.8 & 21.0 & 30.4 \\
        SSUL \cite{cha2021ssul} (NeurIPS21)  & 39.9 & 17.4 & 32.5 & 40.2 & 18.8 & 33.1 & 41.3 & 18.0 & 33.6 & 48.4 & 20.2 & 29.6 \\
        RCIL \cite{zhang2022representation} (CVPR22) & 38.5 & 11.5 & 29.6 & 39.3 & 17.7 & 32.1 & 42.3 & 18.8 & 34.5 & 48.3 & 24.6 & 32.5 \\
        MicroSeg \cite{zhang2022mining} (NeurIPS22)  & 40.4 & 20.5 & 33.8 & 41.5 & 21.6 & 34.9 & 40.2 & 18.8 & 33.1 & 48.6 & 24.8 & 32.9 \\
        EWF \cite{xiao2023endpoints} (CVPR23) & 41.4 & 13.4 & 32.1 & 41.5 & 16.3 & 33.2 & 41.2 & 21.3 & 34.6 & - & - & - \\
        \midrule
        \rowcolor{Gray}
        ENDING (Ours) & {\bf 41.8} & {\bf 21.7} & {\bf 35.1} & {\bf 42.3} & {\bf 22.1} & {\bf 35.6} & {\bf 43.0} & {\bf 22.6} & {\bf 36.2} & {\bf 49.3} & {\bf 25.5} & {\bf 33.6} \\
        \bottomrule
    \end{tabular}}
    \caption{Quantitative comparison on ADE20K. The results indicate that our ENDING yields the state-of-the-art performance in all cases. The best performance is in {\bf bold}.}
    \label{tab:ade_main}
\end{table*}

\subsection{Results on PASCAL VOC 2012}
Table \ref{tab:voc_main} shows the results on PASCAL VOC 2012.
Overall, our method achieves new state-of-the-art performance across all scenarios, affirming that the proposed evolving knowledge mining significantly enhances the model's capacity for incremental learning.
We also make several other observations.
First, our method performs well in challenging scenarios, where there are few old classes and many incremental steps. 
For instance, in the {\bf 2-2 (10 steps)} incremental set-up, where only 2 classes are provided in the $1$-st step, our method effectively preserves performance on old classes without forgetting. 
Simultaneously, by dynamically utilizing diverse features, our method surpasses others in performance on new classes, exhibiting a 13.3\% mIoU gain over the second-best method. 
Second, unlike other methods, our method tends to excel in both preserving old knowledge and exploring new knowledge. 
For example, it consistently achieves the highest averaged mIoU across all classes compared to other approaches, underscoring the efficacy of evolving multi-grained knowledge mining.

Moreover, Figure \ref{fig:visualization} shows the qualitative comparison, which is consistent with the quantitative results.
Interestingly, we found that our method tends to segment the target objects with fine edges while other methods either segment part of the object, \egno, only part of the sofa is segmented in first two rows, or mis-segment some unrelated objects (\egno, the middle region of the cabinet is mis-classified as train).

\subsection{Results on ADE20K}
In Table \ref{tab:ade_main}, we present a comprehensive comparison of existing methods on ADE20K.
Our method again yields the state-of-the-art performance on this challenging dataset with more classes.
We also found that our ENDING outperforms other methods in all cases, indicating its superior ability to address more challenging scenarios.

\subsection{Further Analysis}

\begin{figure}[ht]
  \centering
  \begin{subfigure}[b]{0.235\textwidth}
    \includegraphics[width=\textwidth]{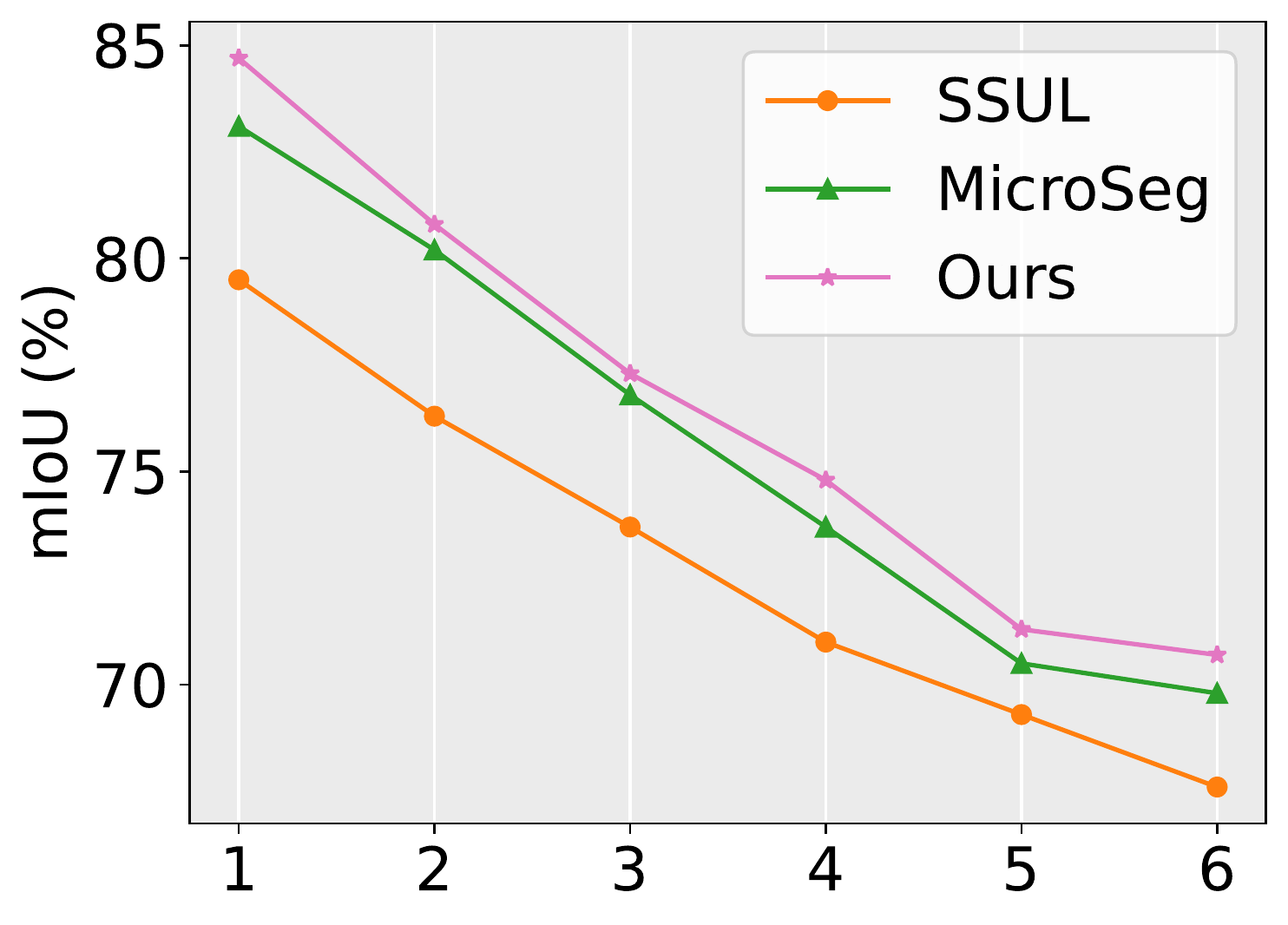}
    \caption{{\bf 15-1 (6 steps)}}
    \label{fig:15-1}
  \end{subfigure}
  \hfill
  \begin{subfigure}[b]{0.235\textwidth}
    \includegraphics[width=\textwidth]{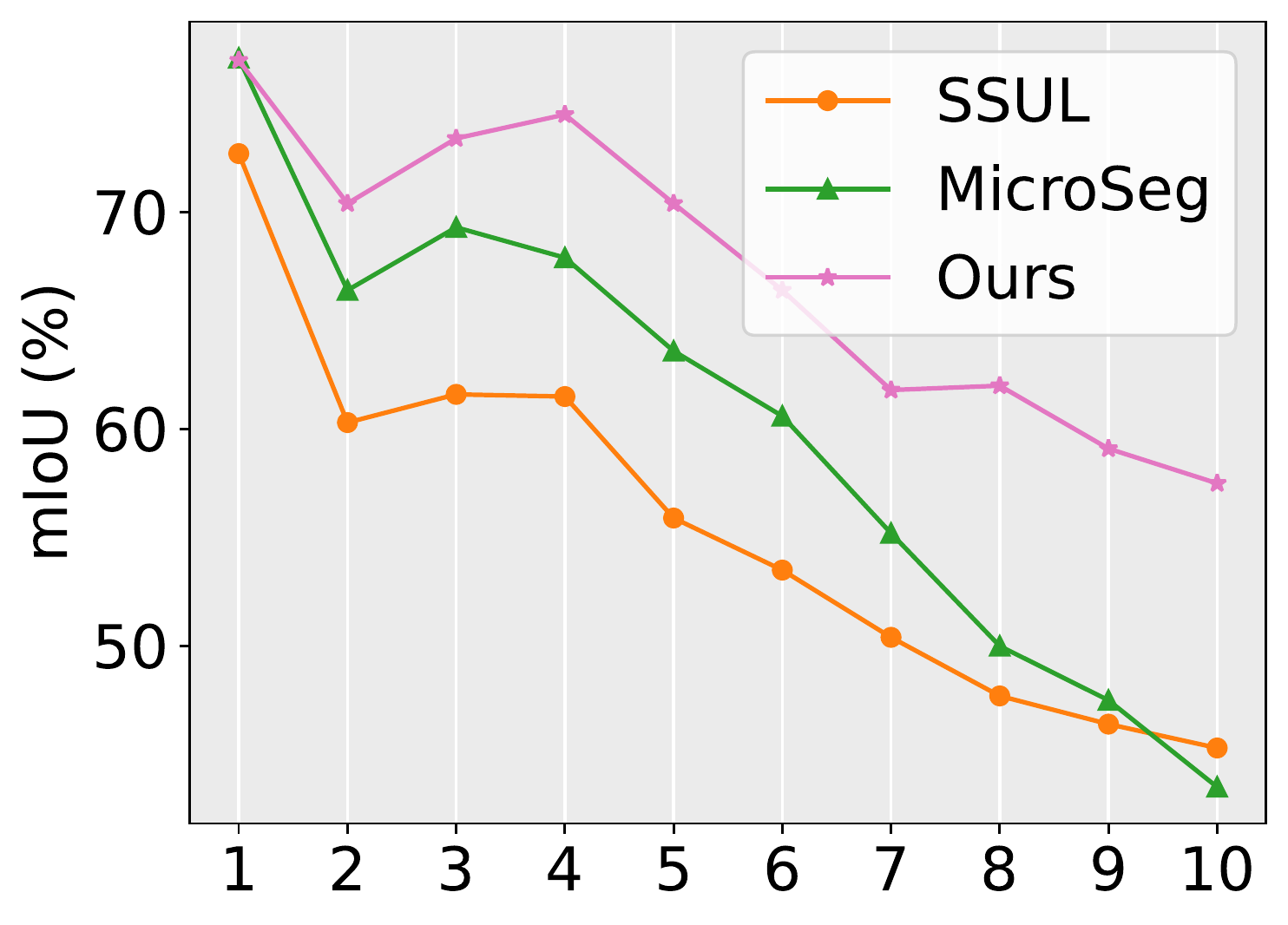}
    \caption{{\bf 2-2 (10 steps)}}
    \label{fig:2-2}
  \end{subfigure}
  \caption{Step-wise mIoU for two set-ups.}
  \label{fig:step}
\end{figure}

\paragraph{Step-wise mIoU}
Figure \ref{fig:step} shows the step-wise mIoU under scenarios {\bf 15-1 (6 steps)} and {\bf 2-2 (10 steps)}, from which we can see that our method consistently yields the best performance against the previous SOTA methods.
This can be attributed to the strong capability of our ENDING in exploring new knowledge without forgetting the old concepts.
Specifically, in the challenging {\bf 2-2 (10 steps)} step-up with more incremental steps, our method initially matches MicroSeg \cite{zhang2022mining} in the $1$-st step, and consistently maintains performance on the seen classes while adapting to new classes. 
This stands in contrast to MicroSeg, which gradually forgets the old knowledge across steps.

\paragraph{Module-wise Ablation Study}
Table \ref{tab:component} presents the ablation study of the two components: evolving fusion and semantic enhancement. 
In summary, both components contribute significantly to the enhancement of the baseline method. 
Notably, the improvement is more pronounced when the initial class number is small, \egno, in the {\bf 2-2 (10 steps)} scenario. 
It is worth highlighting that the performance improvement extends to both old and new classes, underscoring the complementary effect of utilizing both components together.
\begin{table}[ht]
    \centering
    \adjustbox{max width=0.38\textwidth}{
    \begin{tabular}{cc|ccc|ccc}
    \toprule
         \multirow{2}{*}{EF} & \multirow{2}{*}{SE} & \multicolumn{3}{c}{{\bf 15-5 (2 steps)}} & \multicolumn{3}{c}{{\bf 2-2 (10 steps)}} \\
          & & 0-15 & 16-20 & all & 0-2 & 3-20 & all \\
         \midrule
         & & 80.4 & 52.8 & 73.8 & 61.4 & 40.6 & 43.5 \\
         $\checkmark$ & & 80.7 & 54.5 & 74.5 & 63.8 & 49.6 & 51.6 \\
         \rowcolor{Gray}
         $\checkmark$ & $\checkmark$ & {\bf 81.5} & {\bf 56.1} & {\bf 75.5} & {\bf 67.3} & {\bf 55.8} & {\bf 57.4} \\
    \bottomrule
    \end{tabular}}
    \caption{The ablation study on two components: evolving fusion (EF) and semantic enhancement (SE).}
    \label{tab:component}
\end{table}

\paragraph{The Effect of Replaying}
Following MicroSeg \cite{zhang2022mining}, we also verify the effectiveness of the proposed method when an extra memory bank (fixed size=100 \cite{zhang2022mining,cha2021ssul}) for replaying is used, as shown in Table \ref{tab:replay}.
We observed that our ENDING continues to benefit from replaying and outperforms other replaying-based methods by large margins.
\begin{table}[ht]
    \centering
    \adjustbox{max width=0.43\textwidth}{
    \begin{tabular}{l|ccc|ccc}
    \toprule
         \multirow{2}{*}{Method} & \multicolumn{3}{c}{{\bf 15-5 (2 steps)}} & \multicolumn{3}{c}{{\bf 2-2 (10 steps)}} \\
          & 0-15 & 16-20 & all & 0-15 & 16-20 & all \\
         \midrule
         SSUL-M \cite{cha2021ssul} & 78.4 & 55.8 & 73.0 & 58.8 & 45.8 & 47.6 \\
         MicroSeg-M \cite{zhang2022mining} & 82.0 & 59.2 & 76.6 & 60.0 & 50.9 & 52.2 \\
         \rowcolor{Gray}
         ENDING-M & {\bf 82.9} & {\bf 65.5} & {\bf 78.8} & {\bf 66.5} & {\bf 60.3} & {\bf 61.2} \\
    \bottomrule
    \end{tabular}}
    \caption{The effect of replaying on PASCAL VOC 2012.}
    \label{tab:replay}
\end{table}

\paragraph{Evaluation in the Disjoint Set-up}\label{sec:disjoint}
Some past works \cite{douillard2021plop,cermelli2020modeling,zhang2022representation} also evaluate their methods in the {\it disjoint} set-up , which is less realistic as it assumes that all future classes are known and {\it not} shown in the background of the current step.
For a comprehensive comparison, we also verify the effectiveness of our method in this set-up (Table \ref{tab:disjoint}), from which we found the proposed ENDING can consistently improve the baselines, often by large margins.
\begin{table}[ht]
    \centering
    \adjustbox{max width=0.43\textwidth}{
    \begin{tabular}{l|ccc|ccc}
    \toprule
         \multirow{2}{*}{Method} & \multicolumn{3}{c}{{\bf 15-5 (2 steps)}} & \multicolumn{3}{c}{{\bf 15-1 (6 steps)}} \\
          & 0-15 & 16-20 & all & 0-15 & 16-20 & all \\
         \midrule
         LwF-MC \cite{li2017learning} & 67.2 & 41.2 & 60.7 & 4.5 & 7.0 & 5.2 \\
         ILT \cite{michieli2019incremental} & 63.2 & 39.5 & 57.3 & 3.7 & 5.7 & 4.2 \\
         MiB \cite{cermelli2020modeling} & 71.8 & 43.3 & 64.7 & 46.2 & 12.9 & 37.9 \\
         SDR \cite{michieli2021continual} & 74.6 & 44.1 & 67.3 & 59.4 & 14.3 & 48.7 \\
         PLOP \cite{douillard2021plop} & 71.0 & 42.8 & 64.3 & 57.9 & 13.7 & 46.5 \\
         SSUL \cite{cha2021ssul} & 76.4 & {\bf 45.6} & 69.1 & 74.0 & {\bf 32.2} & 64.0 \\
         RCIL \cite{zhang2022representation} & 75.0 & 42.8 & 67.3 & 66.1 & 18.2 & 54.7 \\
         MicroSeg \cite{zhang2022mining} & 77.4 & 43.4 & 69.3 & 73.7 & 24.1 & 61.9 \\
         \midrule
         \rowcolor{Gray}
         ENDING (Ours) & {\bf 78.6} & 45.1 & {\bf 70.6} & {\bf 76.0} & 31.0 & {\bf 65.2} \\
    \bottomrule
    \end{tabular}}
    \caption{The evaluation in the {\it disjoint} set-up on PASCAL VOC 2012.}
    \label{tab:disjoint}
\end{table}

\section{Conclusion}
To our knowledge, this study is the first to explore the efficient multi-grained knowledge reuse by aggregating multi-level feature when freezing the heavy backbone for CISS.
We have empirically shown that leveraging multi-level feature boost model's capability on incremental learning and proposed a novel method, evolving knowledge mining, to dynamically extract multi-grained knowledge, facilitating both the preservation of old knowledge and the exploration of new knowledge.
Experiments have demonstrated the SOTA performance of our method on widely used benchmarks.
{
    \small
    \bibliographystyle{ieeenat_fullname}
    \bibliography{main}
}

\clearpage
\setcounter{page}{1}
\maketitlesupplementary

\begin{figure*}[ht]
    \centering
    \includegraphics[width=\textwidth]{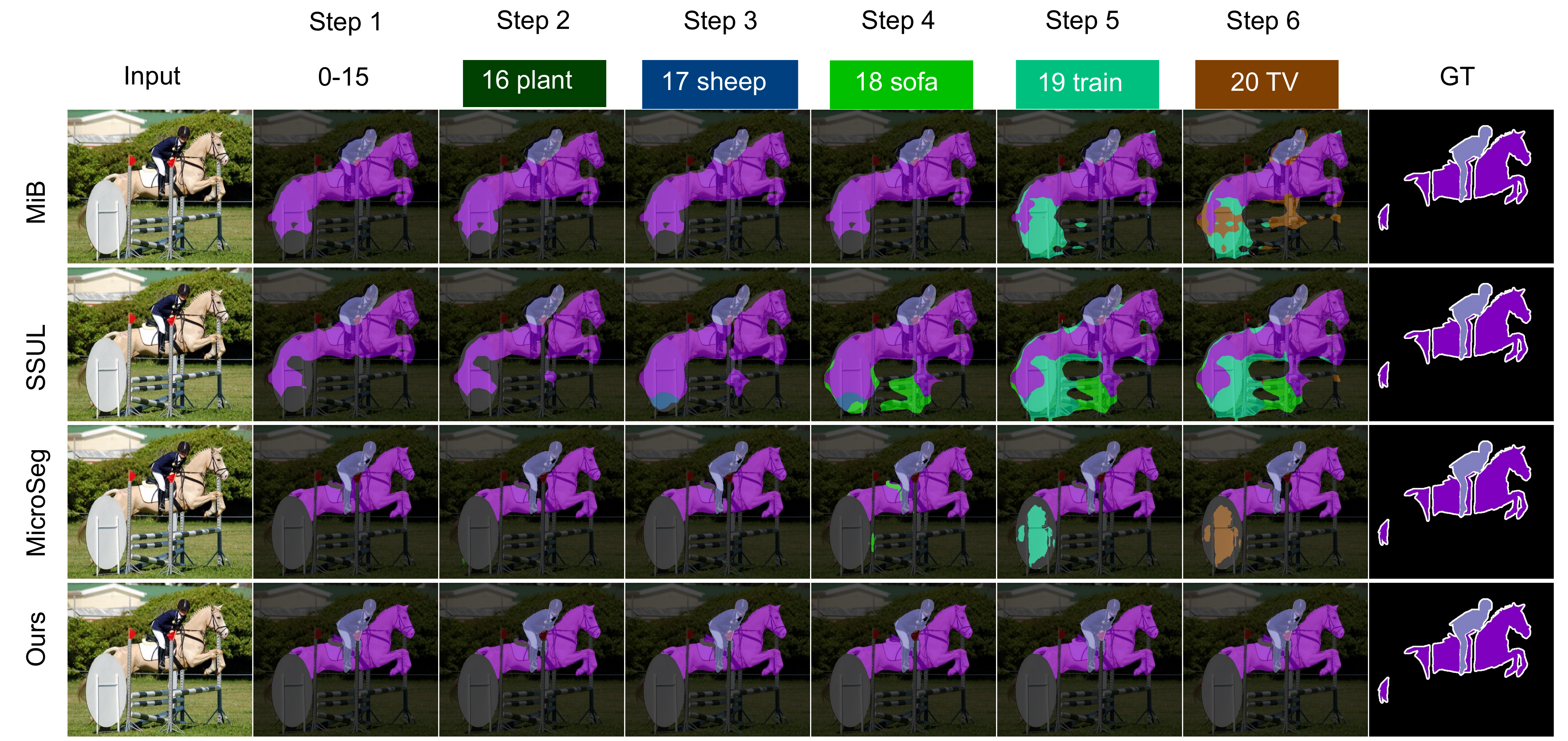}
    \caption{Qualitative comparison with state-of-the-art methods on {\bf 15-1 (6 steps)} setting on PASCAL VOC 2012. Zoom in for details.}
    \label{fig:visual_1}
\end{figure*}

\begin{figure*}[ht]
    \centering
    \includegraphics[width=\textwidth]{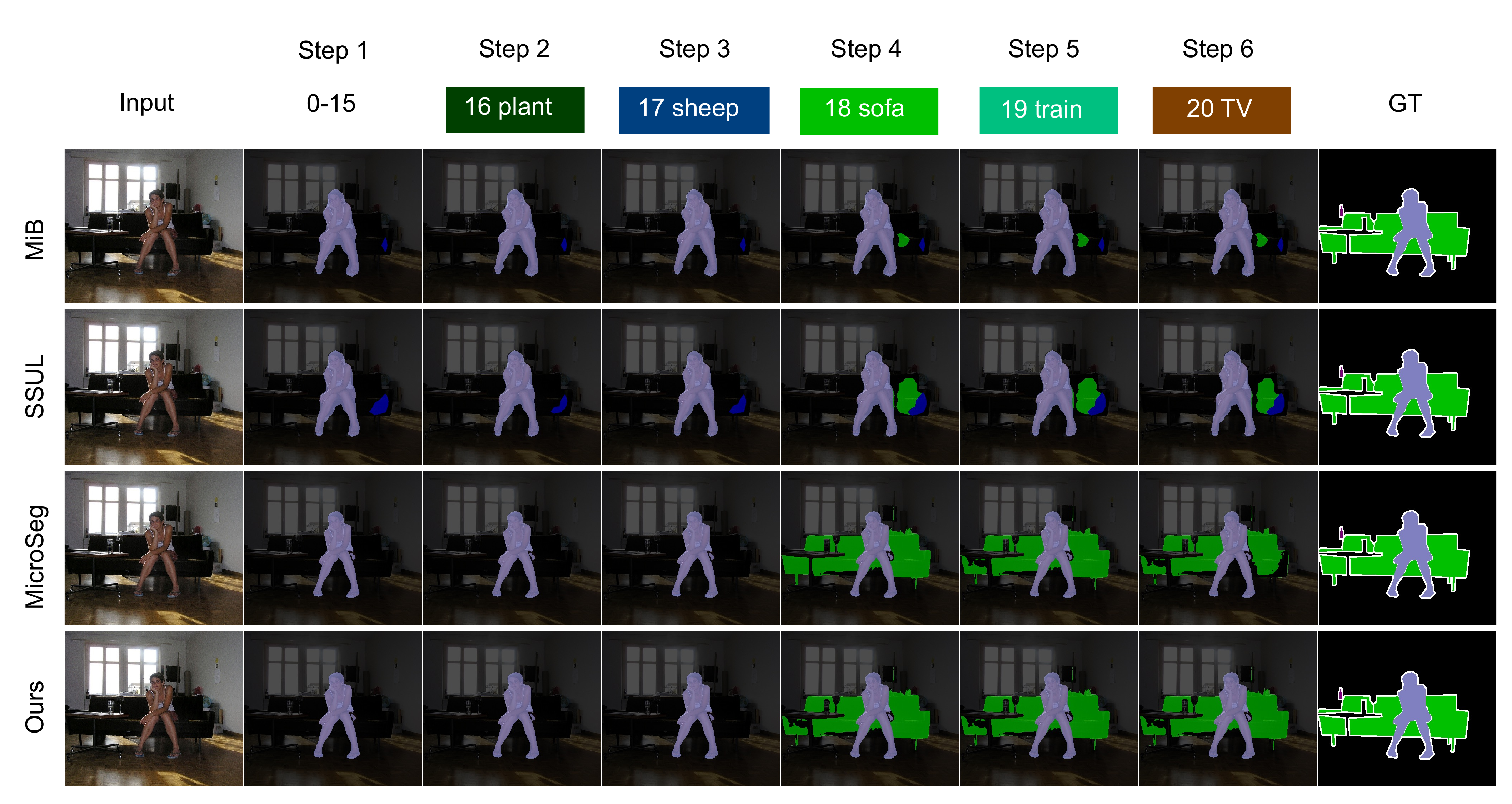}
    \caption{Qualitative comparison with state-of-the-art methods on {\bf 15-1 (6 steps)} setting on PASCAL VOC 2012. Zoom in for details.}
    \label{fig:visual_2}
\end{figure*}

We provide additional content to enhance the understanding of our method, which is listed as follows:
\begin{itemize}
    \item Section \ref{sec:archi} offers the detailed achitucture of our meta-net, which is used for personalized network generation.
    \item Section \ref{sec:exp} shows more experimental results, including the full compared results of replaying-based methods, more qualitative results and failure analysis.
    \item Section \ref{sec:broader} presents the broader impact of our method, including some insightful findings and the significance of our work to the community.
    \item Section \ref{sec:limit} shows the limitations of our method, inspiring the future works.
\end{itemize}

\section{Detailed Architecture}
\label{sec:archi}
We present the detailed architecture of the meta-net employed in ENDING, as illustrated in Table \ref{tab:archi}. 
This meta-net is composed of three sub-nets, each designed to generate personalized networks for one low-level feature. 
Each sub-net takes the corresponding high-level feature $f_4 \in C \times H \times W$ (the output of the ASPP module) as the input and consists of a 2-layer MLP with one activation hidden layer.
To obtain the input for each sub-net, global average pooling is applied to the high-level feature $f_4$. 
The resulting output feature of each sub-net is then reformulated to serve as the weights for a 1 $\times$ 1 Conv layer. 
These Conv layers are instrumental in extracting knowledge from the low-level features. 
The dynamic generation of these Conv layers, conditioned on the high-level feature, empowers ENDING to be class-agnostic. 
This property contributes to its superior performance in incremental learning scenarios without forgetting.
It is noteworthy that the meta-net in ENDING demonstrates remarkable parameter efficiency. 
The total number of parameters in the meta-net is a mere 0.36M, constituting only 0.54\% of the parameters in the entire model.

\begin{table}[ht]
    \centering
    \adjustbox{max width=0.48\textwidth}{
    \begin{tabular}{l|ccc}
    \toprule
         Meta-net & Sub-net $\Theta_1$ & Sub-net $\Theta_2$ & Sub-net $\Theta_3$ \\
         \midrule
        Layer 1 & Linear(256, 4) & Linear(256, 4) & Linear(256, 4) \\
        Activation & ReLU & ReLU & ReLU \\
        Layer 2 & Linear(4, 256 $\times$ 48) & Linear(4, 512 $\times$ 48) & Linear(4, 1024 $\times$ 48) \\
    \bottomrule
    \end{tabular}}
    \caption{Detailed architecture of meta-net. Meta-net consists of three sub-nets, each with a 2-layer MLP and activation layer.}
    \label{tab:archi}
\end{table}

\section{More Experiments}
\label{sec:exp}

\subsection{The Effect of Replaying}
We compare our method with state-of-the-art (SOTA) replaying-based methods across all incremental setups on the PASCAL VOC 2012 dataset, as shown in Table \ref{tab:full_replay}.
Overall, the results consistently reveal that our approach, evolving knowledge mining (ENDING), achieves a new SOTA performance across all scenarios. 
This underscores the superior capabilities of ENDING in incremental learning, emphasizing its efficacy in evolving knowledge mining and adept reuse across diverse features.
Several other observations can be made.
First, we found that our ENDING-M tends to performs much better on challenging scenarios characterized by a small number of classes in the $1$-st step with many incremental steps.
This is consistent to the observation in ENDING.
For instance, ENDING-M outperforms the second best method by 9.0\% in {\bf 2-2 (10 steps)} and 5.0\% in {\bf 5-3 (6 steps)} set-ups, respectively.
Second, with the extra memory used for replaying, our method demonstrates consistently better performance on the old classes.
This is attributed to the excellent ability of our ENDING to leverage additional information for enhanced old knowledge preservation.

\begin{table*}[ht]
    \adjustbox{max width=\textwidth}{
    \centering
    \begin{tabular}{l|ccc|ccc|ccc|ccc|ccc|ccc}
         \toprule
         \multirow{2}{*}{Method} & \multicolumn{3}{c}{{\bf 10-1 (11 steps)}} & \multicolumn{3}{c}{{\bf 2-2 (10 steps)}} & \multicolumn{3}{c}{{\bf 15-1 (6 steps)}} & \multicolumn{3}{c}{{\bf 5-3 (6 steps)}} & \multicolumn{3}{c}{{\bf 19-1 (2 steps)}} & \multicolumn{3}{c}{{\bf 15-5 (2 steps)}} \\
         & 0-10 & 11-20 & all & 0-2 & 3-20 & all & 0-15 & 16-20 & all & 0-5 & 6-20 & all & 0-19 & 20 & all & 0-15 & 16-20 & all \\
         \midrule
         SSUL-M \cite{cha2021ssul} (NeurIPS21)  & 74.0 & 53.2 & 64.1 & 58.8 & 45.8 & 47.6 & 78.4 & 49.0 & 71.4 & 71.3 & 53.2 & 58.4 & 77.8 & 49.8 & 76.5 & 78.4 & 55.8 & 73.0 \\
         MicroSeg-M \cite{zhang2022mining} (NeurIPS22)  & 77.2 & 57.2 & 67.7 & 60.0 & 50.9 & 52.2 & \textbf{81.3} & 52.5 & 74.4 & 74.8 & 60.5 & 64.6 & 79.3 & \textbf{62.9} & 78.5 & 82.0 & 59.2 & 76.6 \\
         \midrule
         \rowcolor{Gray}
         ENDING-M (Ours) & {\bf 79.2} & {\bf 59.4} & {\bf 69.8} & {\bf 66.5} & \textbf{60.3} & \textbf{61.2} & \textbf{81.3} & \textbf{55.6} & \textbf{75.2} & \textbf{78.5} & \textbf{66.0} & \textbf{69.6} & \textbf{80.8} & 62.5 & \textbf{79.9} & \textbf{82.9} & \textbf{65.5} & \textbf{78.8} \\
         \bottomrule
    \end{tabular}}
    \caption{Comparison of replaying-based methods on PASCAL VOC 2012. The best performance is in {\bf bold}.}
    \label{tab:full_replay}
\end{table*}

\subsection{Qualitative Results and Failure Analysis}
\begin{figure*}[ht]
    \centering
    \includegraphics[width=\textwidth]{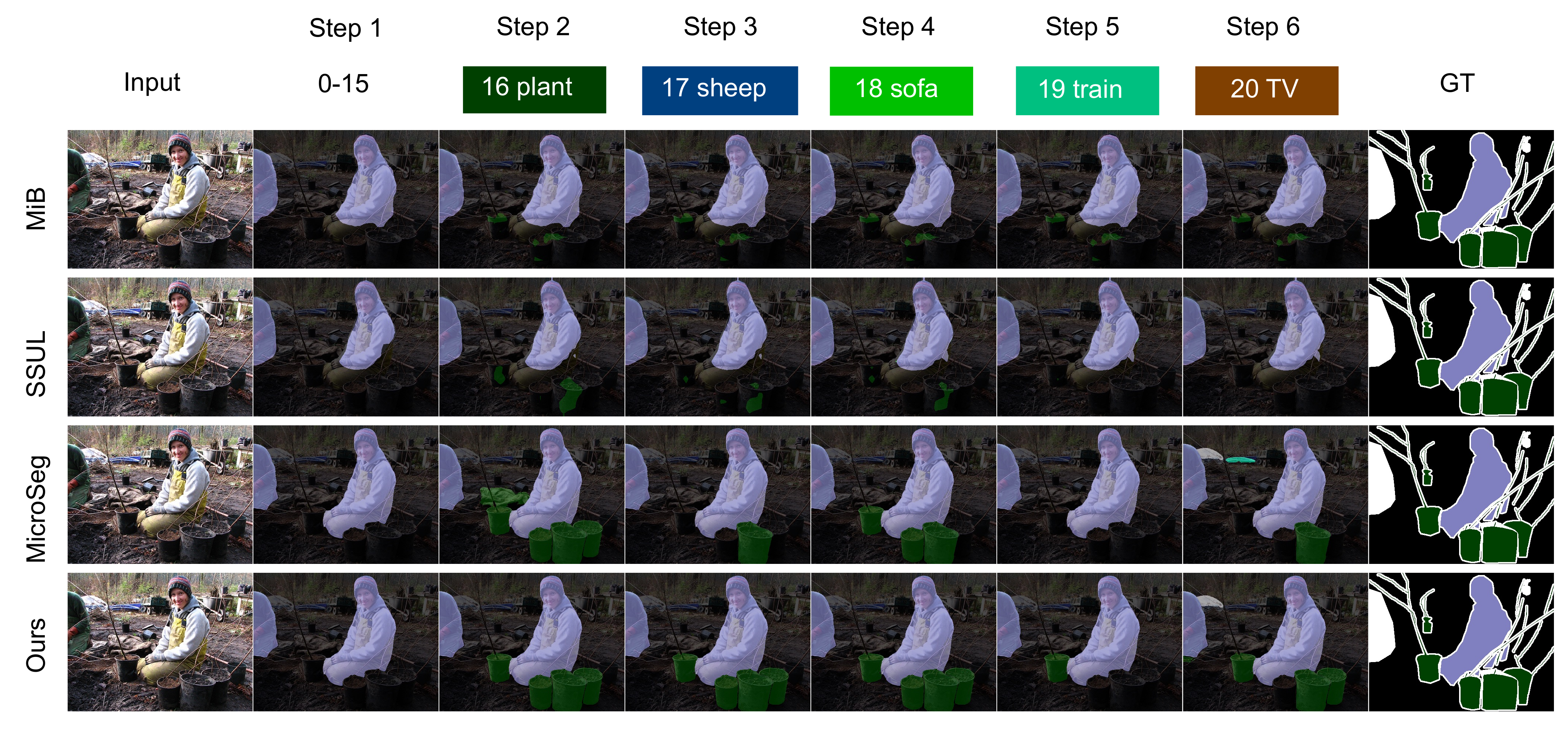}
    \caption{Qualitative comparison with state-of-the-art methods on {\bf 15-1 (6 steps)} setting on PASCAL VOC 2012. Zoom in for details.}
    \label{fig:visual_3}
\end{figure*}

\begin{figure*}[ht]
    \centering
    \includegraphics[width=\textwidth]{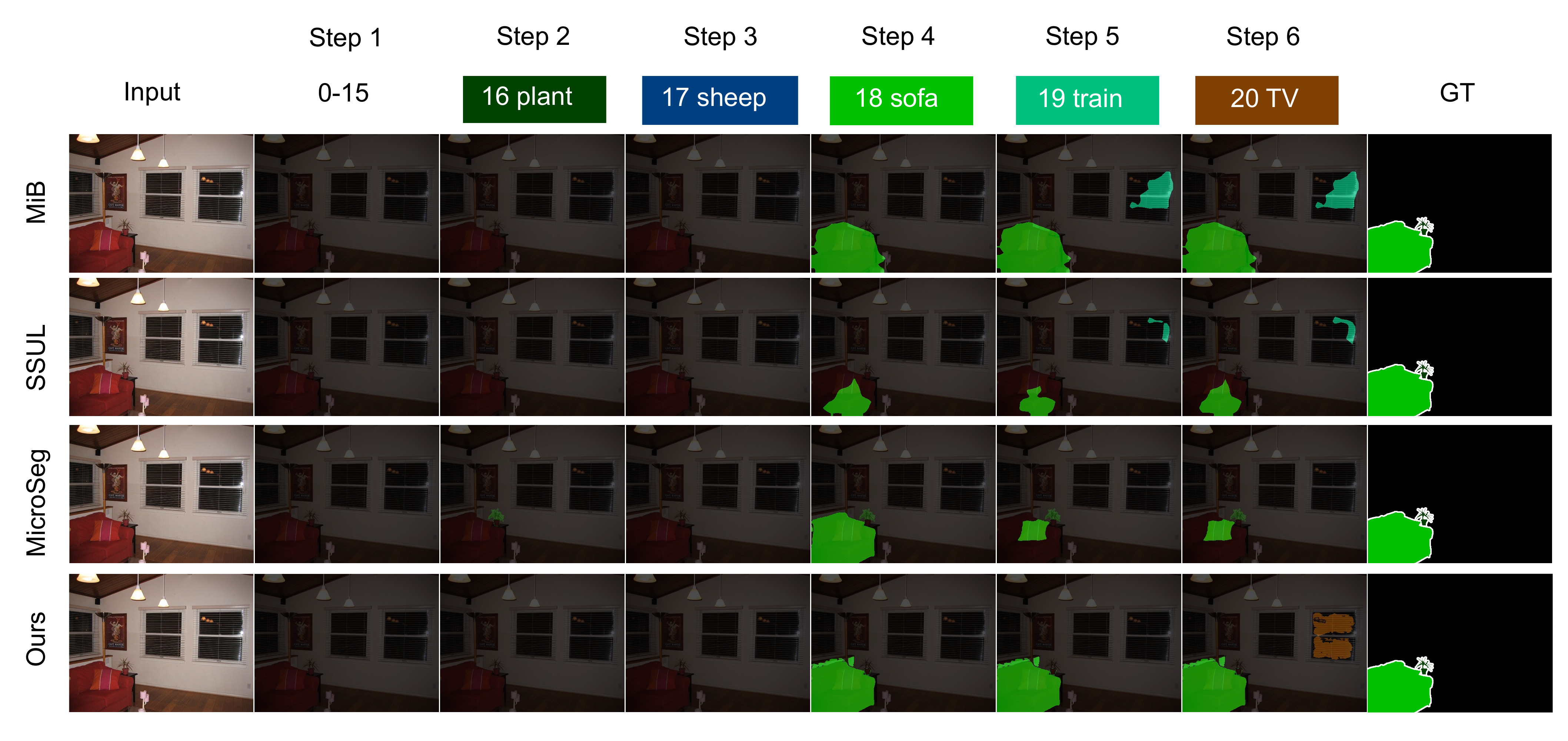}
    \caption{Qualitative comparison with state-of-the-art methods on {\bf 15-1 (6 steps)} setting on PASCAL VOC 2012. Zoom in for details.}
    \label{fig:visual_4}
\end{figure*}

We provide more qualitative results in Figure \ref{fig:visual_1} to \ref{fig:visual_4}.
Overall, our method can consistently produce superior visual results than other existing methods, which is in line with the quantitative results shown in the main text.
In particular, both MiB \cite{cermelli2020modeling} and SSUL \cite{cha2021ssul} exhibit pronounced visual artifacts. 
For instance, as shown in Figure \ref{fig:visual_1}, a board is mis-segmented as part of a horse in SSUL, while MiB displays mis-classification of a board as part of a train.
MicroSeg \cite{zhang2022mining}, on the other hand, tends to classify the board into classes introduced in the current incremental step (\egno, train in the $5$-th step and TV in the last step). 
In contrast, our method excels in preserving old knowledge, avoiding misclassifications into unrelated classes.
Figure \ref{fig:visual_2} further showcases superior capabilities in exploring new knowledge. 
Notably, it accurately captures the shape and structure of the sofa, achieving precise segmentation. 
In contrast, other methods often struggle, segmenting only a portion of the sofa.

Despite achieving superior results overall, as illustrated in Figures \ref{fig:visual_3} and \ref{fig:visual_4}, our ENDING does exhibit some failure cases. 
One notable pattern involves mis-segmentation, where an object is incorrectly assigned to a class with similar appearance and shape. 
For example, a car at a distance is mis-classified as part of a train (Figure \ref{fig:visual_3}, last row) and a window is incorrectly labeled as a TV monitor (Figure \ref{fig:visual_4}, last row).
This observation can serve as valuable insight for future research endeavors.

\section{Broader Impact}
\label{sec:broader}
The findings of this study present a pioneering exploration into the realm of efficient multi-grained knowledge reuse, specifically through the aggregation of multi-level features while maintaining the integrity of a frozen, heavy backbone for class incremental semantic segmentation (CISS). 
This marks a significant advancement in our understanding and application of incremental learning.
The empirical evidence presented in this study underscores the transformative impact of leveraging multi-level features. 
This strategic utilization not only enhances the model's capacity for incremental learning but also introduces a novel approach termed ``evolving knowledge mining''. 
This method dynamically extracts multi-grained knowledge, thereby offering a holistic framework that harmonizes the preservation of existing knowledge with the exploration of new knowledge.
Demonstrating SOTA performance on established benchmarks, this research sets a new standard, offering immediate improvements in model capabilities and paving the way for future innovations in multi-level feature aggregation and incremental learning methodologies.

\section{Limitations}
\label{sec:limit}
We acknowledge that there are multiple ways to design the feature aggregation module, and this work only explores one possibility.
Moving forward, we aim to investigate other potential methods for designing the feature aggregation module. 
We believe that exploring various approaches will allow us to optimize the system's performance and adaptability. 
However, it's important to note that the effectiveness of these methods may vary depending on the specific use case and data characteristics. 
Therefore, further research and experimentation are necessary to identify the most suitable design for different scenarios.


\end{document}